\documentclass[10pt,twocolumn,letterpaper]{article}

\usepackage{iccv}
\usepackage{times}
\usepackage{epsfig}
\usepackage{graphicx}
\usepackage{amsmath}
\usepackage{amssymb}
\usepackage{multirow}
\usepackage{rotating}
\usepackage{booktabs}
\usepackage{adjustbox}
\usepackage[utf8]{inputenc}

\usepackage[pagebackref=true,breaklinks=true,letterpaper=true,colorlinks,bookmarks=false]{hyperref}

\iccvfinalcopy 

\def\iccvPaperID{####} 
\def\httilde{\mbox{\tt\raisebox{-.5ex}{\symbol{126}}}}

\ificcvfinal\fi
\begin{document}

\title{Temporal Accumulative Features for Sign Language Recognition}

\author{Ahmet Alp Kındıroğlu, Oğulcan Özdemir, Lale Akarun\\ 
Computer Engineering Department,\\
Boğaziçi University, Istanbul, Turkey\\
{\tt\small \{alp.kindiroglu,ogulcan.ozdemir,akarun\}@boun.edu.tr}
}

\maketitle
\thispagestyle{empty}

\begin{abstract}
In this paper, we propose a set of features called temporal accumulative features (TAF) for representing and recognizing isolated sign language gestures. By incorporating sign language specific constructs to better represent the unique linguistic characteristic of sign language videos, we have devised an efficient and fast SLR method for recognizing isolated sign language gestures. The proposed method is an HSV based accumulative video representation where keyframes based on the linguistic movement-hold model are represented by different colors. We also incorporate hand shape information and using a small scale convolutional neural network, demonstrate that sequential modeling of accumulative features for linguistic subunits improves upon baseline classification results.      
\end{abstract}

\section{Introduction}
Sign languages are a system of visual communication used by deaf communities around the world. Sign language users also called signers, make use of hand gestures, upper body movements, and facial expressions to form signs and convey meaning visually to other signers. Similar to spoken languages, sign languages have rules that determine their grammar and phonology, allowing users to fluently express themselves with a corpus of thousands of words from simple commands to complex and abstract sentences.

Due to its inherent complex nature and challenges present in its capture, automatic sign language recognition (SLR) has been an active topic of research in the computer vision community for the last 30 years. It is essentially a video classification problem where the task is to find the correct class label by observing the sequence of frames that belong to the video. Due to the complex nature of Sign Language grammars, the problem is tackled in several different levels of complexity. In this study, we are interested in the isolated SLR problem, in which the goal is to perform recognition in a controlled dataset, where disjoint signs are temporally annotated marking their beginning and end locations. In these types of problems, the challenge lies in successfully distinguishing signs among each other, while making sure variations in the performance of the same sign by different users can be recognized as similar. While there exist numerous studies on isolated sign language recognition, the current state of the art in the SLR field makes use of deep learning techniques developed for the action recognition domain. 

Action recognition is an active research field in video recognition, where the availability of huge datasets such as HMDB \cite{kuehne2011hmdb}, UCF-101 \cite{soomro2012ucf101} and Kinetics \cite{kay2017kinetics} have actively driven researchers in academia and industry to push the state of the art. Methods developed in action recognition such as multi-stream networks and 3D convolutional models require large datasets to train models. In the problem of human action recognition, there are several important sources of information such as context, human poses, interactions, and movements. Methods such as I3D exhaustively capture information from those sources to achieve state of the art performance \cite{carreira2017quo}. However, in a recent study called PoTion \cite{choutas2018potion}, it was demonstrated that incorporating accumulated human joint pose information to I3D improved recognition performance with actions where human movements were crucial for differentiation.

Along with its similarities to  human action and activity recognition, sign language recognition has important differentiating characteristics: In sign language videos shot for automatic Sign Language Recognition (SLR), users perform gestures in front of a camera gesticulating with their hands, upper body and faces. Therefore, only fine differences in human pose, hand shape and facial expressions can be used to differentiate among classes. In addition, different performances by individual signers make this problem tougher by increasing inter-class variability where intra-class variability is already low to begin with. Compared to action and activity recognition where context plays a large role, this brings a requirement for SLR developers to rely solely on the success of human pose capture and representation techniques for distinguishing among classes. In addition, sign language videos tend to follow certain grammatical rules such as the Liddell-Johnson \cite{liddell1989american} grammatical model, which assumes that gestures follow a movement-hold pattern sequentially. 

Due to these differences in the nature of SL, in order to better adapt techniques developed for action recognition to SLR, we should focus on methods capturing human motion, hand shapes and movement trajectories instead of context, interactions and variations in appearance. In addition, utilizing the temporal structure to account for sign linguistic rules such as the movement-hold pattern would be beneficial in achieving better temporal alignment in recognizing complex gestures consisting of multiple subunits. In this study, we propose an approach to adapt a number of techniques proposed for human action recognition to the problem of SLR. 

\begin{figure}[t]
\begin{center}
   \includegraphics[width=1.0\linewidth]{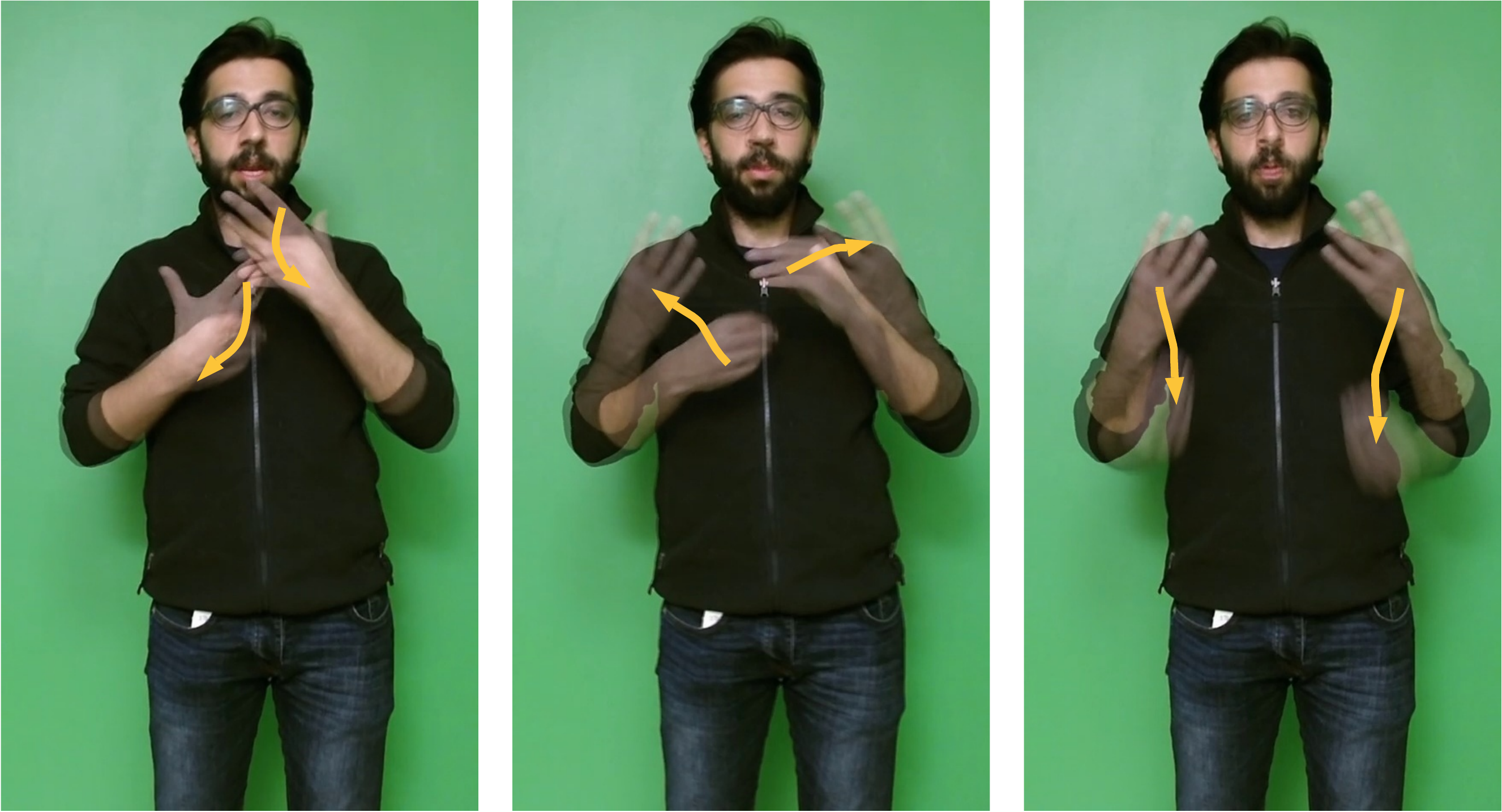}
\end{center}
   \caption{The sign "Crawling" represented using the Liddell-Johnson \cite{liddell1989american} Movement-hold pattern. Most signs follow this grammatical rule, as they may be composed of dynamic and static subunits sequentially following one another.}
\end{figure}

The proposed method uses heat maps obtained from sensor supplied skeleton joints and creates temporally accumulated images from them to construct video level features. These features are represented with a HSV color space based color scheme to represent temporal groupings, temporal location inside the group and heatmap strength. Through keyframe detection, we incorporate dynamic subunits which are represented with different colors based on their order of occurrence. The proposed hand speed and density clustering based approach achieves high quality subunit segmentation. These subunits are coupled with hand shapes representing the hold part of the gestures extracted via an unsupervised hand shape segmentation algorithm. These static subunits are included in the feature set as single frames or several consecutive frames to increase descriptive value. 

The contributions of this paper are as follows: 
\begin{enumerate}
    \item The adaptation of the baseline PoTion method to the sign language recognition domain to perform SLR and finding the optimal model architecture.
    \item A hue based novel temporal colorization scheme that allows distinct representation with arbitrary temporal channels or subunits as well as allowing distinct representations of temporal location and joint heatmap strength.
    \item A static subunit detection method that makes use of local minima in hand speed with density clustering to extract dynamic and static subunits.
    \item Unsupervised hand segmentation to represent hand shapes and utilize these hand shapes as static subunits in the form of keyframes and keyshots with the temporally accumulated features.
\end{enumerate}

\section{Related Work}

Following the success of deep learning approaches on tasks such as image classification and segmentation \cite{krizhevsky2012imagenet, he2016resnet, szegedy2015inception, simonyan15vgg}, recent studies on human action recognition have focused on adapting successful deep image recognition architectures into the video classification domain. While earlier work mostly used 2D CNNs to extend the spatial domain temporally by using different input frames and layer level fusion techniques \cite{karpathy2014large}, later approaches tried to learn videos by exploiting the spatial domain into the temporal domain using 3D CNNs \cite{tran2015learning}, two-stream architectures \cite{simonyan2014two, carreira2017quo, feichtenhofer2016spatiotemporal, feichtenhofer2017multiplier} and recurrent architectures \cite{zhang2016video, ji2017attention, li2018videolstm, donahue2017longterm}. 

Based on the two-stream hypothesis, which supports the idea that human visual cortex consists of two separate pathways; dorsal (motion) and ventral (appearance) streams, \cite{simonyan2014two} used still frames for ventral and stacked optical flow images for dorsal streams, and successfully trained both networks on UCF-101 and HMDB datasets. Although the idea of two-stream networks were thought as networks trained on two classification streams (appearance and motion) \cite{simonyan2014two}, researchers have expanded this idea by adding multiple modalities (e.g. audio spectrograms) for multi-modal action recognition \cite{wu2015fusing}. Recently, I3D  architecture \cite{carreira2017quo} was proposed as an improvement of two-stream networks. In another work, Wang et al. \cite{wang2018tsn} have proposed Temporal Segment Networks (TSNs) with the purpose of solving the long-range temporal limitations of two-stream networks by using temporal sampling. More recently, Choutas et al. \cite{choutas2018potion} proposed PoTion representations for human action recognition. In the method, pose information of individuals in the frame was aggregated over time, and represented as encoding of video clips after a colorization process.   

In the video recognition literature, accumulative features similar to Potion have long been used as simple descriptors to represent videos: Motion History Image (MHI) \cite{bobick1996real} and Motion Energy Image (MEI) \cite{bobick1996appearance} capture energy and motion over the temporal domain.  \cite{wong2005real} uses such a representation to represent the signer's body pose, hand shape and hand movements.   Later on, as these features were found to be insufficient in representing gestures, researchers used more complex features such as Space-Time Interest Points (STIP) \cite{nandakumar2013multi}, Histogram of Oriented Gradients (HOG) \cite{liwicki2009automatic, camgoz2016sign} and Improved Dense Trajectories (IDT) \cite{peng2014action, ozdemir2016siu}.

Similar to human action recognition, sign language recognition was also influenced by the recent developments of deep learning. Pigou et al. explore the idea of merging and forming CNN and RNN based architectures for gesture recognition \cite{pigou2018beyond}, and proposed a CNN based architecture for sign language recognition \cite{pigou2015cnn}. Koller et al. \cite{koller2016deep} proposed a frame-based CNN-HMM architecture for sign language classification problem which was trained on nearly $1$ million hand images. Due to its success on problems such as action recognition, 3D convolutional neural network architectures have also been used for gesture and sign language recognition \cite{molchanov2015hand, camgoz2016using}.  Recently, Camgoz et al. \cite{camgoz2017subunets} used Bi-directional LSTMs to train an end-to-end continuous sign language recognition framework which is based on subunit modelling. Aside from the recognition problems, researchers have also pursued the idea of translating sign language videos into spoken language by adapting neural machine translation approaches into the sign language domain \cite{camgoz2018neural, guo2018translation, stoll18bmvc}.

In the sign language literature, there are various approaches that study subunit based sign language recognition. Similar to the representation of utterances by phonemes in spoken languages, signs may be considered to be composed of subunits concatenated over time. There are studies on the sequential structure of sign languages such as the Stokoe notation \cite{stokoe1980sign} and the works of Liddell and Johnson \cite{liddell1989american}. In applying phonetic modeling to automatic recognition, there are several studies. As there is a lack of subunit corpus and lexicons available for most sign languages, majority of the methods focus on unsupervised learning from data. Some of the notable ones include \cite{fang2004novel}, where the authors use single states in an HMM to cluster dynamic subunits and generate a subunit lexicon. In \cite{kong2010sign}, a rule based phoneme extraction method is used. One study, that makes use of the movement-hold model to construct an end to end sign language recognition model is \cite{theodorakis2014dynamic}. In this study, a phonetic modeling approach for unsupervised dynamic-static subunit extraction is proposed as well as sign subunit modeling with HMMs. 

In sign languages like Turkish sign language, where a lexicon of available subunits is not well defined, researchers have often made use of data driven subunit representation methods. Currently, there exists several hand datasets that can be leveraged for either hand keypoints or segmented hand masks. Deep learning methods have greatly improved the state of the art in image segmentation. Initial deep learning methods such as Fully Convolutional Networks \cite{shelhamer2017fcn} and UNet \cite{ronneberger205unet} have been greatly improved. Current state of the art models in the field include popular approaches such as Deeplab v3+ \cite{chen2017rethinking}, PSPNet, neural architecture search generated Auto-Deeplab and Densely Connected Atrous Spatial Pyramid Pooling (DenseASPP) \cite{yang2018denseaspp}. In this study, we made use of semantic segmentation techniques to represent hand shapes.

\section{Sign Language Recognition Using Temporal Accumulative Features}

Sign languages rely on a sequence of body part configurations to convey a message. To decipher this message, body part detection methods are commonly used. Deep learning methods that solely rely on skeletal information or those that combine skeletal information with visual information have achieved top performances in recognizing signs in large sign language datasets. In our case, the targeted Bosphorus Sign Dataset (Section \ref{sec:bsign}) \cite{camgoz2016bosphorussign} contains high quality upper body joint coordinate annotations and high definition RGB images making the use of accumulative joint information in combination with hand shapes a powerful representation candidate. 

\begin{figure}[t]
\begin{center}
   \includegraphics[width=1.0\linewidth]{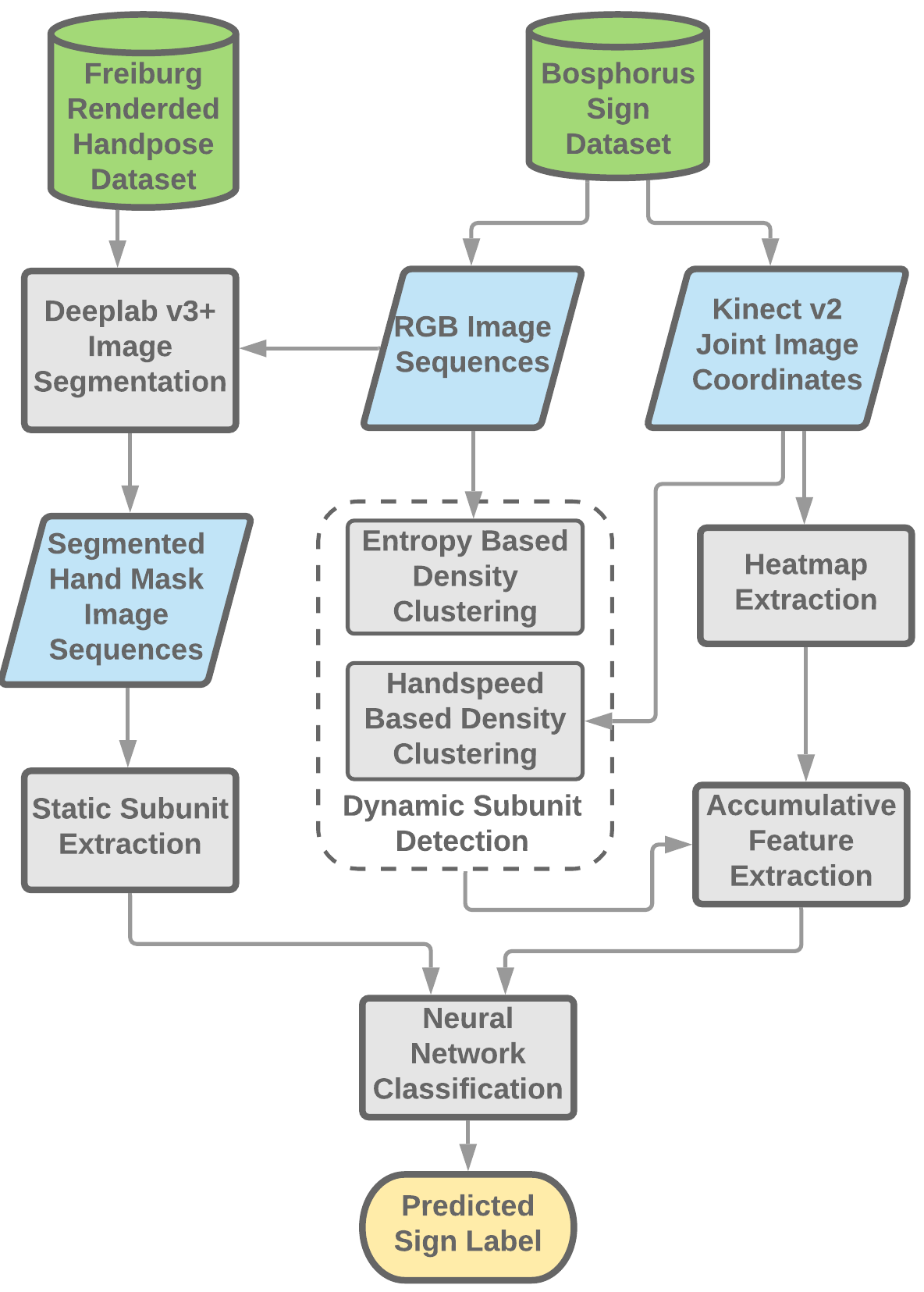}
\end{center}
   \caption{Overview of Proposed Framework.}
\label{fig:pipeline}
\end{figure}

An overview of the proposed framework is presented in Figure \ref{fig:pipeline}. The system employs   accumulative features coupled with subunit representations, the details of which are discussed in the following subsections, to classify signs.

\subsection{Temporal Accumulative Features}
\label{sec:acc}
Temporal Accumulative Features (TAF) is a novel Sign language representation method, that captures a signers body movements and hand shape while taking into account the movement and hold structure shared through  sign languages. Accumulative features like PoTion \cite{choutas2018potion} have shown promising performance on action recognition. The proposed method can be regarded as an adaptation of the PoTion method with major modifications to its inputs and temporal representation methods. 

In constructing temporal accumulative features, we first obtain frame level joint coordinate representations in the form of heatmaps. These heatmaps are aggregated for each video to obtain a temporal representation of the movement. Two different colorization strategies are used to represent sign language gestures in the temporal domain: Linear and subunit-based. Finally, the constructed features are inputs to a Convolutional Neural Network for sign level classification, and a class label of the sign is obtained.

\subsubsection{Obtaining joint heatmaps for TAF}

In order to represent signers upper body configurations, we have made use of the 3D human joint coordinates detected by the Kinect-v2 camera. In the absence of 3D skeleton information, 2D skeleton extraction methods may be employed to extract the skeleton, as in \cite{choutas2018potion}.

We generated  heatmap images using the skeleton joints and assuming Gaussian uncertainties on the joint locations. Of the $25$ Kinect joints, experimentation showed that using only heatmaps of 10 joints belonging to the left and right upper arms, elbows, wrists, hand tips and thumbs and an additional background joint heatmap containing representations of all joints was sufficient. We have scaled all the heatmaps to $116 \times 64$ pixels to achieve better performance comparison with the original PoTion method \cite{choutas2018potion}.

\subsubsection{Accumulating joint heatmaps over time}
\label{sec:jointheatmap}
Prior to obtaining frame level heatmaps, we construct sign level features by creating images containing the aggregation of these heatmaps over time, similar to Motion History Images (MHI) \cite{bobick1996real}. While colorizing these gestures, we use several different temporal and colorization strategies. 

In the isolated sign language dataset, each video consists of $T$ frames. Our baseline temporal strategy involves dividing these $T$ frames into $C$ consecutive groups of equal length, and representing them in separate channels. We also explore the use of more meaningful temporal divisions along temporal subunit boundaries in Section \ref{sec:subunit}. 

\begin{figure*}[t]
\begin{center}
   \includegraphics[width=1\linewidth]{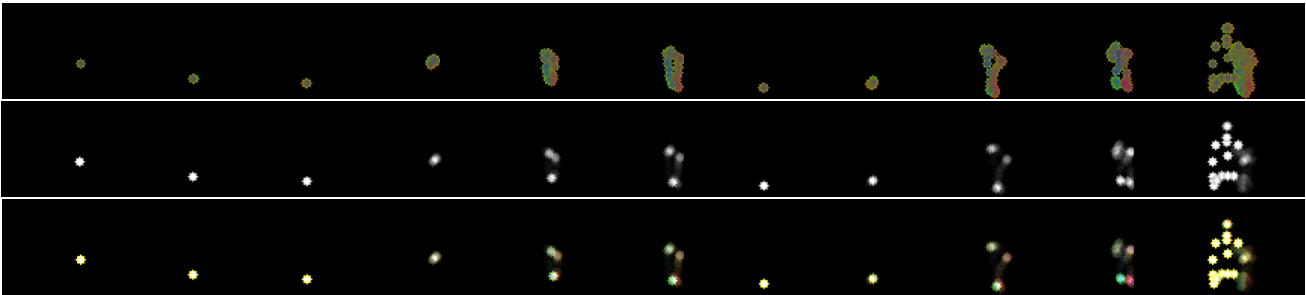}
\end{center}
   \caption{Hue based temporal accumulative feature representation for 11 joints. The representations in the top row are obtained by normalizing the bottom row with the middle row.}
\label{fig:cnn}
\end{figure*}

As a method for differentiating gestures temporally, we use two colorization schemes. The first one is the original colorization scheme proposed in \cite{choutas2018potion}. In this scheme, heatmap images are normalized to the $[0-1]$ range for each joint. The heatmaps are then multiplied by a time based function to be represented in $C$ separate channels, in a $H\times W\times C$ sized image. The images are summed over time and divided by the maximum value of each pixel over all channels to normalize the temporal images with respect to the time spent at a fixed location. In the end, seven $H\times W\times C$ images for each joint are concatenated and used as a feature vector.

As an alternative to the proposed baseline method, a hue based coloring scheme for representing temporal information is defined. Instead of combining heatmap strength and time passed in one channel by multiplying them, we represent them by a HSV color channel image. In this image, the hue channel represents subunits (either $C$ equidistant consecutive temporal segments between frames $0-T$ or detected distinct dynamic keyframes); the saturation channel represents the time passed with respect to the current temporal unit; and the value channel represents the heatmap strength. In this manner, an interval of 8 frames can be represented in 3 channels using 8 different linearly spaced colors from the hue space.

In order to increase the representation power of subunits, a sequence-based variation of the accumulative features was implemented. In this version, each dynamic subunit is represented as an individual gesture. The TAF representation is then calculated by concatenating the feature vectors of all subunits; thus, increasing the total number of channels while decreasing the number of frames in each channel. 

\subsection{Dynamic Subunit Detection and Representation with TAF}
\label{sec:subunit}
In the literature, keyframe, keyshot and video summarization approaches are often used as a means for data abstraction. In this problem, we employ keyframe extraction techniques for linguistic sign language subunit segmentation, which aims to improve the temporal representation of data. 

In keyframe detection our goal is to find a set of frames where motion slows down and changes direction. Given a sign video consisting of a sequence of frames \{$1:T$\}, we propose 3 different methods to designate sign linguistically informative frames as keyframes. 

In the first heuristic, we use handspeed as a feature for detecting keyframes. Frame differences in Kinect world coordinates for the left and right hands are calculated and normalized. Minima locations are marked and a heuristic threshold based on hand speed is used to select the most desirable keyframes. The number of selected keyframes can be limited by choosing the $K$ slowest minima locations for both hands or all keyframes greater than a threshold can be accepted.  

In the entropy based keyframe extraction method, we employ a technique which is based on image entropy and density-peak based clustering \cite{tang2019fast}. This approach first extracts image entropy from video frames and maps them to 2D. Local extrema (minima and maxima) points are considered as most descriptive points of image entropy. Finally, keyframes are extracted by clustering these local extrema using Density clustering \cite{rodriguez2014density}.

A final method is proposed by mixing the best of the first two methods. In the hand speed with density clustering method we propose several improvements. First, we select the dominant hand which is the hand with more motion, and detect keyframes only for the speed changes on that hand. We detect only local minima regions for hand speed and we choose $K+2$ keyframes allowing us to discard the initial and the last keyframes as they often do not contain frames belonging to any active part of the gesture. Finally, density clustering is applied on the hand clusters to find the most descriptive peaks in the movement of the dominant hand. 

\begin{figure}[t]
\begin{center}
   \includegraphics[width=1\linewidth]{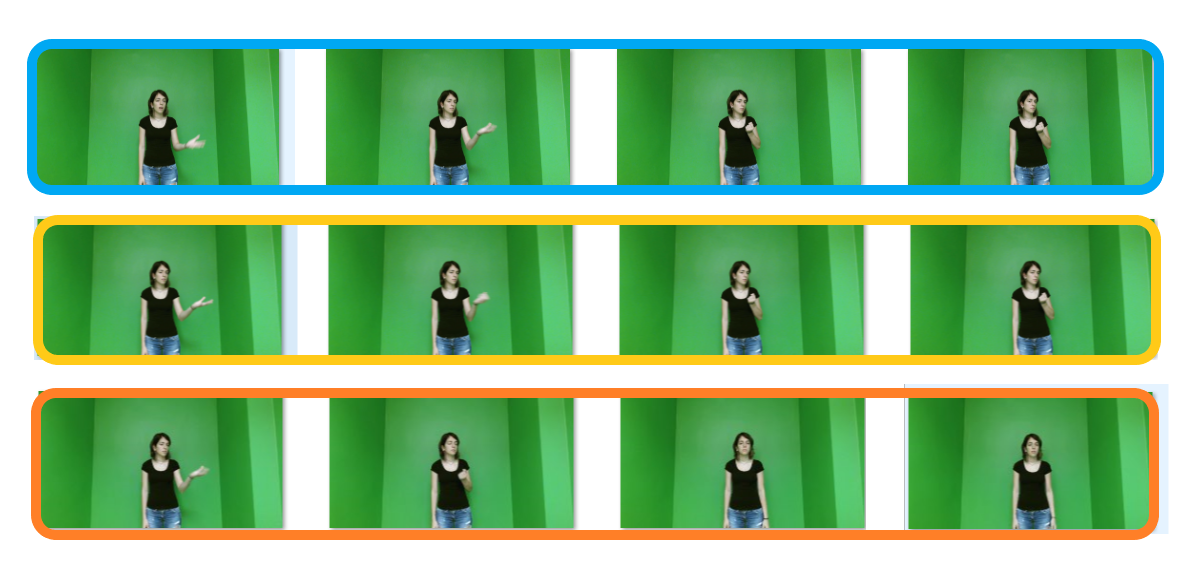}
\end{center}
   \caption{Example selected keyframes using 3 keyframe detection methods. The keyframes detected by the heuristic handspeed method are represented inside blue borders, entropy based keyframes are in yellow and the handspeed based density clustered keyframes are in orange borders.}
\label{fig:onecol}
\end{figure}

In TAF, the representation of the dynamic subunits detected from all three keyframing methods is the same. Given a sign language video, each video is divided into segments separated by the selected keyframes. In the hue based temporal colorization approach, each dynamic segment is represented with a unique hue. For example, for the $n$-th keyframe the chosen color is $n \times 180/(K+1)$ as we do not want the first and the last keyframes to have the same color.

\subsection{Static Subunit Extraction}
\label{sec:Deeplab}
In Turkish Sign Language, there is no widely available lexicon of static handshape based subunits. For that reason, we opted to use data driven subunits that are extracted from our dataset in an unsupervised manner. Since the proposed features do not contain background information, we decided to use segmented handshapes as additional features. 

\begin{figure}[t]
\begin{center}
   \includegraphics[width=0.8\linewidth]{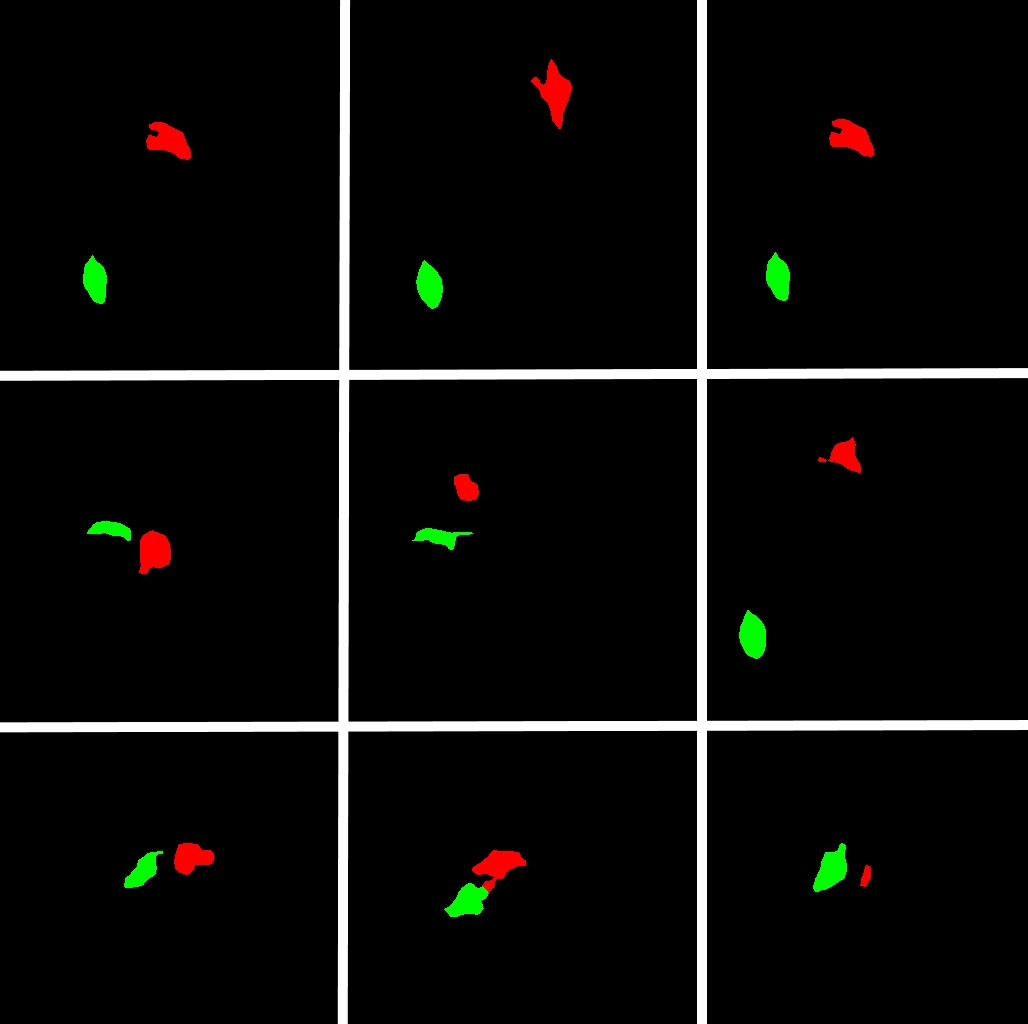}
\end{center}
   \caption{Example hand segmentations from the Bosphorus Sign Dataset. Green masks represent the signers right hands and the red masks represent the signers left hands.}
\label{fig:deeplab}
\end{figure}

In order to train a hand shape segmentation network, we first construct a versatile hand segmentation dataset. In our experiments, we began training a model with the Freibourg Rendered Handpose Dataset \cite{zimmermann2017learning}. The dataset contains $41.258$ training images with their hand joints annotated for segmentation. As the recognition performance of the models trained on this dataset were not satisfactory by themselves, we included the segmented hand images from the HGR dataset \cite{Kawulok2014EURASIP, Nalepa2014BDAS,Grzejszczak2016MTA}. In the ground truth images, we decided on using three classes, namely the left hand, the right hand and background which contained all other pixels. 
We used the Deeplab v3+ model proposed in \cite{chen2018encoder} to perform semantic segmentation. Choosing $1\%$ of images for validation, we achieved mean Intersection over Union (mIoU) scores of \httilde{$78\%$} for the left hand and  \httilde{$83\%$} for the right hand. Example images of segmentations on the Bosphorus Sign dataset can be seen in Figure \ref{fig:deeplab}

Utilizing these segmentations, we constructed and experimented with several representations to incorporate static hand shape information as static subunits into the temporal accumulative feature representation. We propose two representations named as keyframe and keyshot based static subunits. After detecting keyframe locations using one of the methods described in Section \ref{sec:subunit}, the segmented hand masks belonging to the selected frames are resized, converted to binary masks and added to the TAF representation. In the keyframe approach, we input the segmented mask for each detected keyframe. In the keyshot approach, we add five consecutive frames centered at the keyframe. 

\subsection{CNN Based Classification}
\label{sec:neural}
In this section, we give the details of the convolutional neural network model, which is used to perform isolated sign language classification using TAF representation.

\textbf{CNN architecture.} Due to the similarity between TAF representation and the method proposed in \cite{choutas2018potion}, we used a similar CNN architecture which does not need pre-training because our features are based on sparse heatmaps without texture information. In the architecture that we have used, there are 6 convolutional layers which have kernels with the size of 3, and a fully connected layer just before the softmax classification layer. We also apply global average pooling before the fully connected layer for reducing the number of parameters of our model to prevent overfitting. 

In our architecture, we have 2 blocks of layers, each with 2 convolutional layers of  strides 2 and 1, respectively. After each block, we doubled the number of filters of convolutional layers to reduce the spatial dimensions. We have also applied batch normalization and non-linearity after each convolutional layer. 

As for the inputs of our architecture, we stack Temporal Accumulative Features over selected 10 joints (Section \ref{sec:acc}) and use them as inputs to the network. Specifically, we feed TAF representations with size of $H\times W\times C$ into the network, where $H\times W$ is size of the heatmap and $C$ is number of channels. All representation methods and their input dimensionalities for the proposed neural network are shown in Table \ref{tab:table-input}.

\begin{table}[t]
  \centering
  \begin{adjustbox}{max width=\linewidth}
    \begin{tabular}{lrrr}
    Method & \# joints & \# channels & \# static frames \\
    \midrule
    PoTion & 11    & 2c+1  & - \\
    Hue   & 11    & 7     & - \\
    Hue + KF & 11    & 7     & K \\
    Hue + KF + Static keyframes & 11    & 7     & K \\
    Hue + KF + Static keyshot & 11    & 7     & 5K \\
    \bottomrule
    \end{tabular}%
    \end{adjustbox}
    \vspace{3pt}
    \caption{Variations of the Temporal Accumulative Feature methods as inputs in our architecture. ($c$ is the number of color channels, $K$ is the number of static frames and KF is Key Frames). \#joints x \#channels + \#static frames yields the final input dimensionality. } 
  \label{tab:table-input}%
\end{table}%

\textbf{Network training.} Since our features and architecture are much simpler than recent action and sign language recognition models, we train our network from scratch. During the training stage, we first initialize all network layer weights with Xavier initialization \cite{glorot2010init}. In addition, we use Dropout \cite{srivastava14dropout} after convolutional layers, with  probability  $0.5$. We train our network using Adam optimization \cite{diederik2014adam}, with batch size of 32 on a NVIDIA Tesla V100 GPU. 

As most of recent learning approaches, we randomly flipped TAF representations horizontally before feeding them into the network, which increased the performance of our approach drastically. In addition, we have also experimented on channel flipping approach which was proposed in \cite{choutas2018potion}, but the approach had a negative effect on our training, as expected.

\section{Experiments}
In this section, we go over the implementation details, experiment design and present the experimental results. The proposed TAF method has been implemented using the Keras library with a Tensorflow \cite{tensorflow2016} backend. In our experiments, we perform signer independent classification and obtain two recognition metrics: accuracy and top-5 accuracy. We report top-5 accuracy as it demonstrates the potential improvement this method may provide when it is fused with a different type of classifier. 

\subsection{Datasets}
\label{sec:bsign}

The dataset used to validate the results in the paper is the Bosphorus Sign Dataset. The dataset is a publicly open Isolated Sign Language Recognition dataset that is available on request \cite{camgoz2016bosphorussign}. This dataset contains videos of isolated Turkish Sign Language gestures where the gestures begin and end with the rest pose where the signers relax and lower their arms to the sides. Each sign is performed 4-10 times by four to six signers. The signs are recorded from a frontal pose where each signer's sign-space is visible in front of a green background. Two subsets of the Bosphorus Sign Dataset \cite{camgoz2016bosphorussign} were used:  Overlapping Sub-sequences Dataset (OSD) (10 Signs) and the General Subset (174 Signs). In both datasets, user independent tests are performed with $4.839$ training videos and $949$ test videos  on the General subset and $305$ training and $94$ test videos on the OSD. In both datasets user, 4 has been chosen as the test user.   

\subsection{Choosing the best temporal accumulation / coloring strategy}
In this experiment, we first attempt to find a baseline Temporal Accumulative Feature (TAF) method by comparing the two temporal accumulation strategies. The first one is the baseline method where the multi-channel aggregation strategy is identical to the one in the original PoTion paper \cite{choutas2018potion}. The second one is the Hue Temporal Representation Strategy which is introduced in this paper. 

We first experiment with the number of channels parameter determining the number of temporal clusters used to group the $T$ number of frames in a given video. In this experiment, $C$ equidistant consecutive temporal groupings are made. In the original method, these groupings are represented in their respective channels. In the hue based method, these groupings are represented with different hue components in a 3 channel HSV image. The methods are tested on the smaller OSD and the larger Bosphorus Sign General Datasets. 

\begin{table}[t]
  \centering
    \begin{adjustbox}{max width=\linewidth}
    \begin{tabular}{lrrr}
    \multicolumn{1}{r}{} &        & \multicolumn{2}{c}{Dataset} \\
    \toprule
    Aggregation Method & \multicolumn{1}{c}{\# of C} & \multicolumn{1}{l}{OSD} & \multicolumn{1}{l}{General} \\
    \toprule
    \multicolumn{1}{l}{\multirow{7}[0]{*}{Baseline}} & 2      & 98.73  & 72.18 \\
    \multicolumn{1}{c}{} & 3      & 100    & 73.26 \\
    \multicolumn{1}{c}{} & 4      & 100    & 70.71 \\
    \multicolumn{1}{c}{} & 5      & 100    & 64.77 \\
    \multicolumn{1}{c}{} & 6      & 94.93  & 64.46 \\
    \multicolumn{1}{c}{} & 8      & 91.13  & 61.74 \\
    \multicolumn{1}{c}{} & 10     & 82.27  & 60.69 \\
    \midrule
    \multicolumn{1}{l}{\multirow{3}[0]{*}{Hue Temporal Rep.}} & 2      & 100    & 71.12 \\
    \multicolumn{1}{c}{} & 3      & 100    & \textbf{74.52} \\
    \multicolumn{1}{c}{} & 4      & 100    & 72.8 \\
    \bottomrule
    \end{tabular}
    \end{adjustbox}
    \vspace{3pt}
  \caption{Effects of the number of channels (\# of C) parameter on recognition accuracy. }
  \label{tab:table2}
\end{table}

As evident in Table \ref{tab:table2}, using three to four channels is the ideal representation for the baseline PoTion approach. We observe that increasing the number of channels increases the variance in temporal representation. However, increasing the number of channels past a certain point becomes detrimental as similar parts of gestures performed at differing speeds get assigned to different channels. The experiments also showed the better performance of the Hue based temporal representation approach. For the following experiments, we chose this method with three channels as our baseline. In addition, as we achieved 100\% performance on the OSD dataset, we opted not to use it in evaluating further variations of the Hue based Temporal Representation method. 

\subsection{CNN Parameter Optimization for SLR}
\label{sec:cnn-optim}
After establishing the hue based colorization as our baseline method for TAF representations, we ran several experiments with different neural network parameters to find the best architecture. We experimented with four different parameters: namely, dropout, number of  blocks, number of convolution layers in each block and the initial number of filters in the first convolutional layer. We search the parameter space, constraining some parameters at each step  to find the best architecture for this problem.

\begin{table}[t]
  \centering
  \begin{adjustbox}{max width=\linewidth}
    \begin{tabular}{rrrrrr}
    \multicolumn{1}{l}{Dropout} & \multicolumn{1}{l}{Blocks} & \multicolumn{1}{l}{Convs} & \multicolumn{1}{l}{Filters} & \multicolumn{1}{l}{Accuracy} & \multicolumn{1}{l}{Top-5} \\
    \toprule
    0.25   & 1      & 2      & 128    & 70.42  & 94.94 \\
    0.25   & 2      & 2      & 128    & \textbf{78.21}  & \textbf{96.84} \\
    0.25   & 3      & 2      & 128    & 74.52  & 96.21 \\
    \midrule
    0.25    & 2      & 1      & 128    & 68.21  & 93.78 \\
    0.25   & 2      & 2      & 128    & \textbf{78.21}  & \textbf{96.84} \\
    0.5    & 2      & 3      & 128    & \textbf{78.50}  & \textbf{96.84} \\
    0.25   & 2      & 4      & 128    & 76     & 96.21 \\
   \midrule
    0.25   & 2      & 2      & 64     & 75.89  & 96.31 \\
    0.5    & 2      & 2      & 64     & 76.74  & \textbf{97.05} \\
    0.75   & 2      & 2      & 64     & 67.89  & 93.05 \\
    0.25   & 3      & 2      & 256    & 72.01  & 93.8 \\
    0.5    & 3      & 2      & 128    & 73.15  & 96 \\
    0.75   & 3      & 2      & 128    & 70.84  & 93.89 \\
   \midrule
    0.25   & 3      & 2      & 64     & 74.21  & 95.68 \\
    0.75   & 3      & 2      & 128    & 70.84  & 93.89 \\
    0.25   & 3      & 2      & 256    & 72.01  & 93.89 \\
    \bottomrule
    \end{tabular}
    \end{adjustbox}
    \vspace{3pt}
    \caption{Neural network parameter optimization for Hue based temporal colorization without temporal subunits}
  \label{tab:table3}
\end{table}

Examining the results in  Table \ref{tab:table3}, it can be deduced that using approximately six layers (two blocks with three convolutional layers or three blocks with two convolutional layers) yields the best results. We rule out using more than 2 blocks and 3 convolutional layers as those models start to become too complex for the amount of data that we have and start to overfit. Likewise, increasing the dropout value further degrades recognition performance. Of the two models with the top performances, the one with two blocks and 2 convolutional layers per block has fewer parameters and is in practice approximately 1.4 times faster to train. For that reason, we use that model when experimenting with temporal subunits.

\subsection{Detecting and incorporating dynamic subunits}

The effect of the subunit detection method detailed in Section \ref{sec:subunit} is explored in this section. Three temporal subunit detection strategies are employed. These are the handspeed based heuristic keyframe detection method (HS+HEU), entropy based density clustering method (ENT+DC) and the handspeed based density clustering methods (HS+DC). 

\begin{table}[htbp]
  \centering
      \begin{adjustbox}{max width=\linewidth}
    \begin{tabular}{llrrr}
    \multicolumn{2}{p{8.145em}}{ Subunit Method} & \multicolumn{1}{l}{\# of C} & 
    \multicolumn{1}{l}{Accuracy} & \multicolumn{1}{l}{Top-5} \\
    \toprule
    \multicolumn{1}{p{3.645em}}{None} &        & 3      & 78.21  & 96.84 \\
    \multicolumn{1}{p{3.645em}}{HS+HEU} & Fixed Length & 5      & 75.44  & 93.57 \\
    \multicolumn{1}{p{3.645em}}{HS+HEU} & Variable Length & 5      & 74.07  & 92.2 \\
    \midrule
    ENT+DC & Fixed Length & 4      & 76.39  & 95.67 \\
    ENT+DC & Fixed Length & 5      & 78.4   & 96.41 \\
    ENT+DC & Fixed Length & 6      & 79.78  & 97.05 \\
    \midrule
    HS+DC  & Fixed Length & 4      & 79.68  & 96.84 \\
    HS+DC  & Fixed Length & 5      & 80.94  & \textbf{97.47} \\
    HS+DC  & Fixed Length & 6      & \textbf{81.37}  & 97.26 \\
    \midrule
    HS+DC  & Variable Length & 4      & 79.78  & 96.94 \\
    HS+DC  & Variable Length & 5      & \textbf{80.21}  & 96.73 \\
    HS+DC  & Variable Length & 6      & 79.68  & 96.94 \\
    \bottomrule
    \end{tabular}
    \end{adjustbox}
  \vspace{3pt}
  \caption{Comparison of temporal subunit detection strategies}
  \label{tab:table4}
\end{table}

In the first heuristic-handspeed approach, we obtain a fixed set of keyframes with the most likely N candidates. In the second, heuristic-handspeed approach, we select frames that are more likely than a given threshold value as keyframes. This yields variable length keyframes. Visualizing the selected frames, we observe inconsistencies between different variations of the same sign, explaining the  drop in accuracy.

In the entropy based density clustering method, we explored three fixed keyframe sizes of 4,5 and 6. The results improved on the baseline results with no keyframes by $1\%$. However, especially in local maxima, the entropy based method found blurry and highly mobile frames that were undesirable when trying to capture the movements and holds in the sign. Finally, with the proposed handspeed based density clustering approach, we achieved near optimal handspeed minima detection. In terms of the number of keyframes parameter, there are two approaches: fixed length and variable length. We observed that the fixed length approach showed significantly more accuracy then the latter method. 

\subsection{Incorporating static subunits}
Lastly, we turn our attention to static subunits. As described in Section \ref{sec:Deeplab}, we have keyframe and keyshot based representations. We fix all other parameters to the best method presented in Table \ref{tab:table4}, using hue based temporal colorization and handspeed based dynamic subunits with density based clustering. We change keyframe length to observe the effect of adding different number of keyframes per gesture. 

\begin{table}[t]
  \centering
    \begin{adjustbox}{max width=\linewidth}
    \begin{tabular}{p{1.5cm}p{1.5cm}lrr}
    Static Subunit & Keyframe Method & Seq.Rep & \multicolumn{1}{l}{Accuracy} & \multicolumn{1}{l}{Top-5} \\
    \toprule
    None   & FL-6   & Single & 81.37  & 97.26 \\
    \midrule
    Keyframe & FL-6   & Single & 81.47  & 97.15 \\
    \midrule
    Keyshot & FL-4   & Single & 81.58  & 96.94 \\
    Keyshot & FL-6   & Single &  80.10      & 97.05  \\
    \bottomrule
    \end{tabular}
    \end{adjustbox}
    \vspace{3pt}
    \caption{Comparison of incorporating static subunits to hand speed and clustering based dynamic keyframes. FL-\# means \# keyframes were chosen with the fixed length strategy in each video}
  \label{tab:table5}
\end{table}

As can be seen in Table \ref{tab:table5}, the addition of static keyframes brings a marginal increase to the detected keyframes. Using keyshots with a smaller keyframe length parameter further increases performance marginally, raising the top prediction score to $81.58\%$.

\section{Conclusions}
In this work, we introduced Temporal Accumulative Features, which is a pose-based visual representation based on the idea of aggregating joint heatmaps over sign language videos, and applied them to isolated sign classification. We proposed a hue based temporal colorization scheme that allowed distinct representation with arbitrary temporal channels or subunits as well as allowing distinct representations of temporal location and joint heatmap strengths. The hue based temporal colorization scheme achieved 78.50\% accuracy improving over the baseline method by 5\%.

We proposed a static subunit detection method that makes use of local minima in hand speed with density clustering to extract dynamic and static subunits. Utilizing that method with the hue based temporal classification, we further increased our accuracy by 3\% to 81.37\% We developed a handshape based segmentation method using the Deeplab v3+ algorithm and incorporated the segmented handshapes as static subunit features to the temporal accumulative features method. Incorporating static subunits further improved our results to 81.58\%. Overall, our experiments on Bosphorus Sign dataset have shown that TAF representations are an effective method for sign language recognition. 

\textbf{Acknowledgements:} This work was funded by the Turkish ministry of development under the TAM Project \#2007K120610, TUBITAK Project \#117E059 and Bogazici Uni. BAP Project \#14504. The numerical calculations reported in this paper were also partially performed at TUBITAK ULAKBIM, HPAGCC-TRUBA.

{\small
\bibliographystyle{ieee}
\bibliography{references}

\begin{thebibliography}{10}\itemsep=-1pt

\bibitem{tensorflow2016}
M.~Abadi, P.~Barham, J.~Chen, Z.~Chen, A.~Davis, J.~Dean, M.~Devin,
  S.~Ghemawat, G.~Irving, M.~Isard, M.~Kudlur, J.~Levenberg, R.~Monga,
  S.~Moore, D.~G. Murray, B.~Steiner, P.~Tucker, V.~Vasudevan, P.~Warden,
  M.~Wicke, Y.~Yu, and X.~Zheng.
\newblock Tensorflow: A system for large-scale machine learning.
\newblock In {\em 12th USENIX Symposium on Operating Systems Design and
  Implementation (OSDI 16)}, pages 265--283, 2016.

\bibitem{bobick1996appearance}
A.~Bobick and J.~Davis.
\newblock An appearance-based representation of action.
\newblock In {\em ICPR, 1996}, volume~1, pages 307--312, August 1996.

\bibitem{bobick1996real}
A.~Bobick and J.~Davis.
\newblock Real-time recognition of activity using temporal templates.
\newblock In {\em Applications of Computer Vision, 1996. WACV '96., Proceedings
  3rd IEEE Workshop on}, pages 39--42. IEEE, December 1996.

\bibitem{camgoz2016using}
N.~C. Camgoz, S.~Hadfield, O.~Koller, and R.~Bowden.
\newblock Using convolutional 3d neural networks for user-independent
  continuous gesture recognition.
\newblock In {\em ICPR, 2016}, pages 49--54. IEEE, December 2016.

\bibitem{camgoz2017subunets}
N.~C. Camgoz, S.~Hadfield, O.~Koller, and R.~Bowden.
\newblock Subunets: End-to-end hand shape and continuous sign language
  recognition.
\newblock In {\em ICCV, 2017}, 2017.

\bibitem{camgoz2018neural}
N.~C. Camgoz, S.~Hadfield, O.~Koller, H.~Ney, and R.~Bowden.
\newblock Neural sign language translation.
\newblock In {\em CVPR, 2018}, 2018.

\bibitem{camgoz2016sign}
N.~C. Camg{\"o}z, A.~A. K{\i}nd{\i}ro{\u{g}}lu, and L.~Akarun.
\newblock Sign language recognition for assisting the deaf in hospitals.
\newblock In {\em International Workshop on Human Behavior Understanding},
  pages 89--101. Springer, 2016.

\bibitem{camgoz2016bosphorussign}
N.~C. Camg{\"o}z, A.~A. Kindiroglu, S.~Karab{\"u}kl{\"u}, M.~Kelepir, A.~S.
  {\"O}zsoy, and L.~Akarun.
\newblock Bosphorussign: A turkish sign language recognition corpus in health
  and finance domains.
\newblock In {\em LREC}, 2016.

\bibitem{carreira2017quo}
J.~Carreira and A.~Zisserman.
\newblock Quo vadis, action recognition? a new model and the kinetics dataset.
\newblock In {\em CVPR, 2017}, pages 4724--4733. IEEE, 2017.

\bibitem{chen2017rethinking}
L.-C. Chen, G.~Papandreou, F.~Schroff, and H.~Adam.
\newblock Rethinking atrous convolution for semantic image segmentation.
\newblock {\em arXiv preprint arXiv:1706.05587}, 2017.

\bibitem{chen2018encoder}
L.-C. Chen, Y.~Zhu, G.~Papandreou, F.~Schroff, and H.~Adam.
\newblock Encoder-decoder with atrous separable convolution for semantic image
  segmentation.
\newblock In {\em ECCV, 2018}, pages 801--818, 2018.

\bibitem{choutas2018potion}
V.~{Choutas}, P.~{Weinzaepfel}, J.~{Revaud}, and C.~{Schmid}.
\newblock Potion: Pose motion representation for action recognition.
\newblock In {\em CVPR, 2018}, pages 7024--7033, June 2018.

\bibitem{donahue2017longterm}
J.~{Donahue}, L.~A. {Hendricks}, M.~{Rohrbach}, S.~{Venugopalan},
  S.~{Guadarrama}, K.~{Saenko}, and T.~{Darrell}.
\newblock Long-term recurrent convolutional networks for visual recognition and
  description.
\newblock {\em TPAMI}, 39(4):677--691, April 2017.

\bibitem{fang2004novel}
G.~Fang, X.~Gao, W.~Gao, and Y.~Chen.
\newblock A novel approach to automatically extracting basic units from chinese
  sign language.
\newblock In {\em ICPR, 2004}, volume~4, pages 454--457. IEEE, 2004.

\bibitem{feichtenhofer2016spatiotemporal}
C.~Feichtenhofer, A.~Pinz, and R.~Wildes.
\newblock Spatiotemporal residual networks for video action recognition.
\newblock In {\em NIPS}, pages 3468--3476, 2016.

\bibitem{feichtenhofer2017multiplier}
C.~Feichtenhofer, A.~Pinz, and R.~P. Wildes.
\newblock Spatiotemporal multiplier networks for video action recognition.
\newblock In {\em CVPR, 2017}, 2017.

\bibitem{glorot2010init}
X.~Glorot and Y.~Bengio.
\newblock Understanding the difficulty of training deep feedforward neural
  networks.
\newblock In {\em Proceedings of the Thirteenth International Conference on
  Artificial Intelligence and Statistics}, volume~9, pages 249--256. PMLR,
  13--15 May 2010.

\bibitem{Grzejszczak2016MTA}
T.~Grzejszczak, M.~Kawulok, and A.~Galuszka.
\newblock Hand landmarks detection and localization in color images.
\newblock {\em Multimedia Tools and Applications}, 75(23):16363--16387, 2016.

\bibitem{guo2018translation}
D.~Guo, W.~Zhou, H.~Li, and M.~Wang.
\newblock Hierarchical lstm for sign language translation.
\newblock In {\em AAAI Conference on Artificial Intelligence}, 2018.

\bibitem{he2016resnet}
K.~{He}, X.~{Zhang}, S.~{Ren}, and J.~{Sun}.
\newblock Deep residual learning for image recognition.
\newblock In {\em CVPR, 2016}, pages 770--778, June 2016.

\bibitem{ji2017attention}
Z.~Ji, K.~Xiong, Y.~Pang, and X.~Li.
\newblock Video summarization with attention-based encoder-decoder networks.
\newblock {\em CoRR}, abs/1708.09545, 2017.

\bibitem{karpathy2014large}
A.~Karpathy, G.~Toderici, S.~Shetty, T.~Leung, R.~Sukthankar, and L.~Fei-Fei.
\newblock Large-scale video classification with convolutional neural networks.
\newblock In {\em CVPR, 2014}, pages 1725--1732, June 2014.

\bibitem{Kawulok2014EURASIP}
M.~Kawulok, J.~Kawulok, J.~Nalepa, and B.~Smolka.
\newblock Self-adaptive algorithm for segmenting skin regions.
\newblock {\em EURASIP Journal on Advances in Signal Processing},
  2014(170):1--22, 2014.

\bibitem{kay2017kinetics}
W.~Kay, J.~Carreira, K.~Simonyan, B.~Zhang, C.~Hillier, S.~Vijayanarasimhan,
  F.~Viola, T.~Green, T.~Back, P.~Natsev, et~al.
\newblock The kinetics human action video dataset.
\newblock {\em arXiv:1705.06950}, 2017.

\bibitem{diederik2014adam}
D.~P. Kingma and J.~Ba.
\newblock Adam: A method for stochastic optimization.
\newblock {\em CoRR}, abs/1412.6980, 2014.

\bibitem{koller2016deep}
O.~Koller, H.~Ney, and R.~Bowden.
\newblock Deep hand: How to train a cnn on 1 million hand images when your data
  is continuous and weakly labelled.
\newblock In {\em CVPR, 2016}, pages 3793--3802. IEEE, June 2016.

\bibitem{kong2010sign}
W.~Kong and S.~Ranganath.
\newblock Sign language phoneme transcription with rule-based hand trajectory
  segmentation.
\newblock {\em Journal of Signal Processing Systems}, 59(2):211--222, 2010.

\bibitem{krizhevsky2012imagenet}
A.~Krizhevsky, I.~Sutskever, and G.~E. Hinton.
\newblock Imagenet classification with deep convolutional neural networks.
\newblock In {\em NIPS}, pages 1097--1105, 2012.

\bibitem{kuehne2011hmdb}
H.~Kuehne, H.~Jhuang, E.~Garrote, T.~Poggio, and T.~Serre.
\newblock Hmdb: A large video database for human motion recognition.
\newblock In {\em 2011 International Conference on Computer Vision}, pages
  2556--2563. IEEE, November 2011.

\bibitem{li2018videolstm}
Z.~Li, K.~Gavrilyuk, E.~Gavves, M.~Jain, and C.~G.~M. Snoek.
\newblock Videolstm convolves, attends and flows for action recognition.
\newblock {\em Computer Vision and Image Understanding}, 166:41--50, 2018.

\bibitem{liddell1989american}
S.~K. Liddell and R.~E. Johnson.
\newblock American sign language: The phonological base.
\newblock {\em Sign language studies}, 64(1):195--277, 1989.

\bibitem{liwicki2009automatic}
S.~Liwicki and M.~Everingham.
\newblock Automatic recognition of fingerspelled words in british sign
  language.
\newblock In {\em CVPRW, 2009}, pages 50--57, June 2009.

\bibitem{molchanov2015hand}
P.~Molchanov, S.~Gupta, K.~Kim, and J.~Kautz.
\newblock Hand gesture recognition with 3d convolutional neural networks.
\newblock In {\em CVPRW, 2015}, pages 1--7. IEEE, June 2015.

\bibitem{Nalepa2014BDAS}
J.~Nalepa and M.~Kawulok.
\newblock Fast and accurate hand shape classification.
\newblock In S.~Kozielski, D.~Mrozek, P.~Kasprowski, B.~Malysiak-Mrozek, and
  D.~Kostrzewa, editors, {\em Beyond Databases, Architectures, and Structures},
  volume 424 of {\em Communications in Computer and Information Science}, pages
  364--373. Springer, 2014.

\bibitem{nandakumar2013multi}
K.~Nandakumar, K.~W. Wan, S.~M.~A. Chan, W.~Z.~T. Ng, J.~G. Wang, and W.~Y.
  Yau.
\newblock A multi-modal gesture recognition system using audio, video, and
  skeletal joint data.
\newblock In {\em ICMI, 2013}, pages 475--482. ACM, 2013.

\bibitem{ozdemir2016siu}
O.~\"{O}zdemir, N.~C. Camgöz, and L.~Akarun.
\newblock Isolated sign language recognition using improved dense trajectories.
\newblock In {\em 2016 24th Signal Processing and Communication Application
  Conference (SIU)}, pages 1961--1964. IEEE, May 2016.

\bibitem{peng2014action}
X.~Peng, L.~Wang, Z.~Cai, and Y.~Qiao.
\newblock Action and gesture temporal spotting with super vector
  representation.
\newblock In {\em ECCVW, 2014}, pages 518--527. Springer International
  Publishing, 2015.

\bibitem{pigou2015cnn}
L.~Pigou, S.~Dieleman, P.-J. Kindermans, and B.~Schrauwen.
\newblock Sign language recognition using convolutional neural networks.
\newblock In L.~Agapito, M.~M. Bronstein, and C.~Rother, editors, {\em ECCVW,
  2014}, pages 572--578, Cham, 2015. Springer International Publishing.

\bibitem{pigou2018beyond}
L.~Pigou, A.~van~den Oord, S.~Dieleman, M.~Van~Herreweghe, and J.~Dambre.
\newblock Beyond temporal pooling: Recurrence and temporal convolutions for
  gesture recognition in video.
\newblock {\em IJCV, 2018}, 126(2):430--439, April 2018.

\bibitem{rodriguez2014density}
A.~Rodriguez and A.~Laio.
\newblock Clustering by fast search and find of density peaks.
\newblock {\em Science}, 344(6191):1492--1496, 2014.

\bibitem{ronneberger205unet}
O.~Ronneberger, P.Fischer, and T.~Brox.
\newblock U-net: Convolutional networks for biomedical image segmentation.
\newblock In {\em Medical Image Computing and Computer-Assisted Intervention
  (MICCAI)}, volume 9351 of {\em LNCS}, pages 234--241. Springer, 2015.

\bibitem{shelhamer2017fcn}
E.~{Shelhamer}, J.~{Long}, and T.~{Darrell}.
\newblock Fully convolutional networks for semantic segmentation.
\newblock {\em TPAMI}, 39(4):640--651, April 2017.

\bibitem{simonyan2014two}
K.~Simonyan and A.~Zisserman.
\newblock Two-stream convolutional networks for action recognition in videos.
\newblock In {\em NIPS}, volume~1, pages 568--576. MIT Press, 2014.

\bibitem{simonyan15vgg}
K.~Simonyan and A.~Zisserman.
\newblock Very deep convolutional networks for large-scale image recognition.
\newblock In {\em ICLR}, 2015.

\bibitem{soomro2012ucf101}
K.~Soomro, A.~R. Zamir, and M.~Shah.
\newblock Ucf101: A dataset of 101 human actions classes from videos in the
  wild.
\newblock {\em arXiv:1212.0402}, 2012.

\bibitem{srivastava14dropout}
N.~Srivastava, G.~Hinton, A.~Krizhevsky, I.~Sutskever, and R.~Salakhutdinov.
\newblock Dropout: A simple way to prevent neural networks from overfitting.
\newblock {\em JMLR}, 15:1929--1958, 2014.

\bibitem{stokoe1980sign}
W.~C. Stokoe.
\newblock Sign language structure.
\newblock {\em Annual Review of Anthropology}, 9(1):365--390, 1980.

\bibitem{stoll18bmvc}
S.~Stoll, N.~C. Camgoz, S.~Hadfield, and R.~Bowden.
\newblock Sign language production using neural machine translation and
  generative adversarial networks.
\newblock In {\em BMVC, 2018}, Newcastle, UK, 3 -- 6 Sept. 2018. BMVA Press.

\bibitem{szegedy2015inception}
C.~{Szegedy}, P.~{Sermanet}, S.~{Reed}, D.~{Anguelov}, D.~{Erhan},
  V.~{Vanhoucke}, and A.~{Rabinovich}.
\newblock Going deeper with convolutions.
\newblock In {\em CVPR, 2015}, pages 1--9, June 2015.

\bibitem{tang2019fast}
H.~Tang, H.~Liu, W.~Xiao, and N.~Sebe.
\newblock Fast and robust dynamic hand gesture recognition via key frames
  extraction and feature fusion.
\newblock {\em Neurocomputing}, 331:424--433, 2019.

\bibitem{theodorakis2014dynamic}
S.~Theodorakis, V.~Pitsikalis, and P.~Maragos.
\newblock Dynamic--static unsupervised sequentiality, statistical subunits and
  lexicon for sign language recognition.
\newblock {\em Image and Vision Computing}, 32(8):533--549, 2014.

\bibitem{tran2015learning}
D.~Tran, L.~Bourdev, R.~Fergus, L.~Torresani, and M.~Paluri.
\newblock Learning spatiotemporal features with 3d convolutional networks.
\newblock In {\em ICCV, 2015}, pages 4489--4497. IEEE, December 2015.

\bibitem{wang2018tsn}
L.~{Wang}, Y.~{Xiong}, Z.~{Wang}, Y.~{Qiao}, D.~{Lin}, X.~{Tang}, and L.~{Van
  Gool}.
\newblock Temporal segment networks for action recognition in videos.
\newblock {\em TPAMI}, pages 1--1, 2018.

\bibitem{wong2005real}
S.-F. Wong and R.~Cipolla.
\newblock Real-time adaptive hand motion recognition using a sparse bayesian
  classifier.
\newblock In {\em Computer Vision in Human-Computer Interaction}, pages
  170--179. Springer Berlin Heidelberg, 2005.

\bibitem{wu2015fusing}
Z.~Wu, Y.-G. Jiang, X.~Wang, H.~Ye, and X.~Xue.
\newblock Multi-stream multi-class fusion of deep networks for video
  classification.
\newblock In {\em Proceedings of the 2016 ACM on Multimedia Conference}, pages
  791--800. ACM, 2016.

\bibitem{yang2018denseaspp}
M.~Yang, K.~Yu, C.~Zhang, Z.~Li, and K.~Yang.
\newblock Denseaspp for semantic segmentation in street scenes.
\newblock In {\em CVPR, 2018}, pages 3684--3692, 2018.

\bibitem{zhang2016video}
K.~Zhang, W.-L. Chao, F.~Sha, and K.~Grauman.
\newblock Video summarization with long short-term memory.
\newblock In {\em ECCV}. Springer, 2016.

\bibitem{zimmermann2017learning}
C.~Zimmermann and T.~Brox.
\newblock Learning to estimate 3d hand pose from single rgb images.
\newblock In {\em ICCV, 2017}, pages 4903--4911, 2017.

\end{thebibliography}
}

\end{document}